\pdfoutput=1

\documentclass[11pt]{article}

\usepackage{ACL2023}

\usepackage{times}
\usepackage{xcolor}
\usepackage{color}
\usepackage{latexsym}
\usepackage{tikz-dependency}

\usepackage[T1]{fontenc}

\usepackage[utf8]{inputenc}

\usepackage{microtype}
\usepackage{pgfplots}
\usepackage{booktabs}

\usepackage{url}

\usepackage{amsmath}
\usepackage{multirow}
\usepackage{adjustbox}

\usepackage{graphicx}  

\usepackage{inconsolata}

\usepackage{CJKutf8}

\title{Enhancing Few-shot NER with Prompt Ordering based Data Augmentation}

\author{
    \textbf{
        Huiming Wang\thanks{~~Huiming Wang is an intern at DAMO Academy, Alibaba Group.} \textsuperscript{\rm ~1,2}~~~
        Liying Cheng \textsuperscript{\rm 1}~~~
        Wenxuan Zhang \textsuperscript{\rm 1}~~~
        De Wen Soh \textsuperscript{\rm 2}~~~
        Lidong Bing\thanks{$^\dag$ Corresponding author.}$^\dag$\textsuperscript{\rm 1}
    }\\
\textsuperscript{\rm 1}DAMO Academy, Alibaba Group~~
\textsuperscript{\rm 2}Singapore University of Technology and Design\\
{\tt huiming\_wang@mymail.sutd.edu.sg~~~~dewen\_soh@sutd.edu.sg} \\
{\tt\{liying.cheng, saike.zwx, l.bing\}@alibaba-inc.com}
}

\begin{document}
\maketitle
\begin{abstract}
Recently, data augmentation (DA) methods have been proven to be effective for pre-trained language models (PLMs) in low-resource settings, including few-shot named entity recognition (NER).
However, conventional NER DA methods are mostly aimed at sequence labeling models, i.e., token-level classification, and few are compatible with unified autoregressive generation frameworks, which can handle a wider range of NER tasks, such as nested NER. 
Furthermore, these generation frameworks have a strong assumption that the entities will appear in the target sequence with the same \textit{left-to-right} order as the source sequence. In this paper, we claim that there is no need to keep this strict order, and more diversified but reasonable target entity sequences can be provided during the training stage as a novel DA method. Nevertheless, a naive mixture of augmented data can confuse the model since one source sequence will then be paired with different target sequences. Therefore, we propose a simple but effective \textbf{Prompt Ordering based Data Augmentation (PODA)} method to improve the training of unified autoregressive generation frameworks under few-shot NER scenarios. Experimental results on three public NER datasets and further analyses demonstrate the effectiveness of our approach.
\footnote{Our code and data can be found at \href{https://github.com/DAMO-NLP-SG/PODA-NER}{https://github.com/\\DAMO-NLP-SG/PODA-NER}.}
\end{abstract}

\begin{figure*}
  \centering
  \includegraphics[width= 1\linewidth]{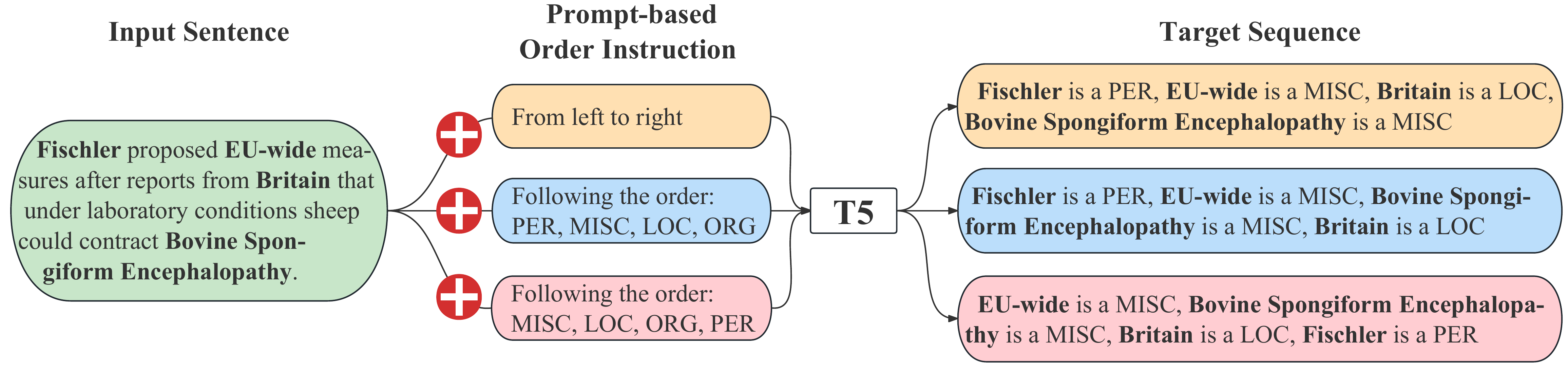}
  \vspace{-2mm}
  \caption{A diagram of our proposed PODA. Every prompt-based order instruction is concatenated with the input sentence as a new source sequence to form the one-to-one mapping with a re-ordered target sequence (rather than following only the \textit{left-to-right} order). Without any modification to the backbone, we can augment the training samples by several times. ``T5'' refers to the backbone of our method.}
  \label{fig:model}
  \vspace{-2mm}
\end{figure*}

\section{Introduction} \label{introduction}
Named entity recognition (NER)~\cite{tjong2003introduction, doddington-etal-2004-automatic-2} has been a long-standing and one of the most important fundamental tasks in natural language processing (NLP). Existing NER models can be divided into three different categories, including sequence labeling methods~\cite{lample2016neural,devlin2019bert}, span-level classification~\cite{wang-lu-2020-two-2, zhong-chen-2021-frustratingly-2} and generation-based methods~\cite{yan-etal-2021-unified-generative-2, lu-etal-2022-unified-2}. However, even with pre-trained language models (PLMs), training these state-of-the-art named entity recognizers requires sufficient training samples, which is in contrast with real-world scenarios, where only small amounts of labeled data are available. This draws our attention to the challenging but practical problem: few-shot NER.


By introducing more samples in the training stage, data augmentation (DA) methods have been proven to be effective solutions in low-resource settings, including few-shot NER~\cite{chen-etal-2020-local-2, zhou-etal-2022-melm-2, chen-etal-2021-data-2}. However, these approaches are mostly designed in a token-level classification style. So they must be combined with different tagging schemes or special-designed structures before they can be applied to other NER subtasks, such as nested NER. On the other hand, generation-based models~\cite{yan-etal-2021-unified-generative-2, Paolini2021StructuredPA} can overcome this limitation by leveraging generative PLMs and introducing a unified tagging strategy. However, few efforts are made on DA over generative PLMs (e.g. BART~\cite{lewis-etal-2020-bart-2}, T5~\cite{Raffel2019ExploringTL}), so limited data resources will lead to weakly fine-tuning of these generation-based methods. Hence, developing a DA approach that can easily be compatible with these generative PLMs would be worthwhile.


Due to the autoregressive decoding of generative PLMs, vanilla generation-based NER methods have a strong assumption that the entities will appear in the target sequence with the same \textit{left-to-right} order as the source sequence. For example, in Figure \ref{fig:model}, there are 4 entities in the input sentence. The prediction of entity \textbf{\textit{EU-wide}} will be strictly after that of entity \textbf{\textit{Fischler}}, following their order of appearing in the input sentence. However, in the NER task, the output entities are essentially forming an unordered set. To mitigate the above-mentioned challenge, significant efforts have been made from different aspects. \citet{Tan2021ASN} proposed a sequence-to-set network and relied mainly on non-autoregressive generation~\cite{gu2018nonautoregressive}. Although they were able to predict the entities as a set, they might suffer from uncertain boundaries, and non-autoregressive generation may also lead to tremendous search space. \citet{zhang-etal-2022-de-2} tried to address this issue still in an autoregressive perspective and constructed augmented samples based on the entities' context and positional orders. However, a simple mixture of these target sequences can confuse the model since there will be several ``gold'' target sequences corresponding to a same source sequence, which will result in a one-to-many mapping problem (also known as \textit{multimodality} problem~\cite{gu2018nonautoregressive}), especially harmful in few-shot NER settings.


In this work, we try to fully utilize the order-agnostic property of NER, and propose a simple but effective \textbf{Prompt Ordering based Data Augmentation (PODA)} method for few-shot NER. In our view, any sequence containing complete information (i.e. every entity's mention and its type) should be regarded reasonable and can serve as an augmented target sequence. With the help of different prompt-based order instructions, we separate the original one-to-many mapping into various one-to-one mappings. As shown in Figure \ref{fig:model}, following a certain entity type permutation like ``PER, MISC, LOC, ORG'', a unique target sequence will then be constructed, and its related source sequence will be the combination of source sentence and the order instruction. In this way, the strict \textit{left-to-right} order does not need to be maintained. 

In summary, our contributions include: (1) We for the first time propose a new data augmentation method which can be uniformly applied over several generative PLMs. Furthermore, we combine our augmented data with prompt-based order instructions to prevent one-to-many mapping problem; (2) Experiments over three benchmark NER datasets, including flat and nested NER, demonstrate the effectiveness of our data augmentation method. Further analyses show the strong generalization ability of our method and validity of our augmented data.


\section{Prompt Ordering based Data Augmentation}
In this section, we firstly introduce the formulation of NER task in a generation style, and then we describe how we construct augmented data without any information from the input sentence. After that, we will illustrate the details of our prompt-based order instructions.

\begin{table*}[ht]
	\centering
	\renewcommand\thetable{1}
	\setlength{\tabcolsep}{6pt} 
	\renewcommand{\arraystretch}{1} 
	\resizebox{0.98\linewidth}{!}{
    \begin{tabular}{ll|cccc}
			\toprule
   \multirow{1}{*}{\textbf{Datasets}}&\multirow{1}{*}{\textbf{Models}} & \textit{K=5} & \textit{K=10}   & \textit{K=20}& \textit{K=50}\\
   \midrule
   \midrule
   \multirow{10}{*}{\textbf{CoNLL-2003}}&T5& 36.77  $\pm$  9.06\:\; &	43.32  $\pm$  3.15\:\;	& 61.76  $\pm$  2.11 &70.35  $\pm$  1.34\\
   &Flan-T5& 44.08  $\pm$  4.96\:\; &	59.35  $\pm$  0.81\:\;	& 67.94  $\pm$  2.86 &72.74  $\pm$  0.92\\
   \cmidrule(lr){2-6}
   &NNShot~\cite{yang-katiyar-2020-simple-2} & 42.31  $\pm$  8.92\:\; & 59.24  $\pm$  11.71 & 66.89  $\pm$  6.09 & 72.63 $\pm$ 3.42\\
   &StructShot~\cite{yang-katiyar-2020-simple-2} & 45.82  $\pm$ 10.30 & 62.37 $\pm$ 10.96 & 69.51 $\pm$ 6.46&74.73 $\pm$ 3.06\\
   &Template-NER~\cite{cui-etal-2021-template-2}& 43.04 $\pm$ 5.15\:\; & 57.86 $\pm$ 5.68\:\;&66.38 $\pm$ 6.09&72.71 $\pm$ 2.13\\
   &Ent-LM~\cite{ma-etal-2022-template-2}& 49.59  $\pm$  8.30\:\; & 64.79 $\pm$ 3.86\:\; & 69.52  $\pm$  4.48 & 73.66 $\pm$ 2.06\\
   &Ent-LM + Struct~\cite{ma-etal-2022-template-2} & 51.32 $\pm$ 7.67\:\; & \textbf{66.86 $\pm$ 3.01\:\;} & 71.23  $\pm$  3.91 & 74.80 $\pm$ 1.87\\
   \cmidrule(lr){2-6}
   &PODA (T5)& \textbf{59.31  $\pm$  1.85\:\;}&	65.54  $\pm$  1.18\:\;&	\textbf{71.68  $\pm$  0.80}&	\textbf{75.66  $\pm$  0.23}\\
   &PODA (Flan-T5)& 58.07 $\pm$  1.28\:\;& 64.79  $\pm$  1.32\:\;& 69.37  $\pm$  1.25& 73.09  $\pm$  1.03\\
   
   \midrule
   \midrule
   \multirow{10}{*}{\textbf{MIT-Movie}}&T5& 53.17  $\pm$  4.05\:\;&	62.96  $\pm$  1.17\:\;&	68.14  $\pm$  0.86&	72.49  $\pm$  0.57\\
   &Flan-T5& 55.74  $\pm$  3.16\:\; &	62.52  $\pm$  0.81\:\;	& 68.59  $\pm$  0.79 &73.09  $\pm$  0.43\\
   \cmidrule(lr){2-6}
   &NNShot~\cite{yang-katiyar-2020-simple-2} & 38.97  $\pm$  5.54\:\; & 50.47  $\pm$  6.09\:\; & 58.94  $\pm$  3.47 & 71.17 $\pm$ 2.85\\
   &StructShot~\cite{yang-katiyar-2020-simple-2} & 41.60  $\pm$  8.97\:\; & 53.19  $\pm$  5.52\:\; & 61.42  $\pm$  2.98&72.07  $\pm$  6.41\\
   &Template-NER~\cite{cui-etal-2021-template-2} &45.97 $\pm$  3.86\:\;& 49.30  $\pm$ 3.35\:\; &59.09  $\pm$ 0.35& 65.13  $\pm$ 0.17\\
   &Ent-LM~\cite{ma-etal-2022-template-2}& 46.62  $\pm$  9.46\:\; & 57.31 $\pm$ 3.72\:\; & 62.36  $\pm$  4.14 & 71.93 $\pm$ 1.68\\
   &Ent-LM + Struct~\cite{ma-etal-2022-template-2} & 49.15 $\pm$ 8.91\:\; & 59.21 $\pm$ 3.96\:\; & 63.85  $\pm$  3.70 & 72.99 $\pm$ 1.80\\
   \cmidrule(lr){2-6}
   &PODA (T5)& 62.14  $\pm$  1.19\:\;&	\textbf{66.62  $\pm$  0.76\:\;}&	\textbf{70.03  $\pm$  0.38}&	\textbf{74.08  $\pm$  0.41} \\
   &PODA (Flan-T5)& \textbf{62.86  $\pm$  1.25\:\;}&	65.50  $\pm$  1.22\:\;&	68.81  $\pm$  0.18&	73.02  $\pm$  0.56\\
   \midrule
   \midrule
   \multirow{5}{*}{\textbf{ACE-2005}}&T5& 25.71  $\pm$  7.41\:\;&	28.78  $\pm$  7.02\:\;&	34.47  $\pm$  2.57&	43.10  $\pm$  0.40\\
   &Flan-T5& 28.30  $\pm$  3.51\:\;&	35.26  $\pm$  1.24\:\;&	39.05  $\pm$  1.20&	43.82  $\pm$  1.00\\
   \cmidrule(lr){2-6}
   &PODA (T5)& 33.42  $\pm$  1.31\:\;&	38.73  $\pm$  2.60\:\;&	\textbf{42.22  $\pm$  1.75}&	44.85  $\pm$  1.19 \\
   &PODA (Flan-T5)& \textbf{36.44  $\pm$  1.49\:\;}&\textbf{40.45  $\pm$  1.76\:\;}&	41.36  $\pm$  2.35&	\textbf{45.22  $\pm$  0.57}\\
			\bottomrule
		\end{tabular}
}
	\vspace{-2mm}
	\caption{The performance on three datasets with different (K=5, 10, 20, 50) few-shot settings. We report the mean and deviation results over 3 different splits for each cell.}
	\label{tab:mainresult}
	\vspace{-3mm}
\end{table*}

\subsection{Formulation}
The NER tasks aim at detecting all the spans that can represent entities within a given sentence $X=[x_1, x_2, ..., x_n]$, where $n$ is the sentence length. The entities in sentence $X$ form the corresponding target sequence $Y$. The $i$-th entity $y_i \in Y$ can be represented as a tuple $y_i = (s_i, t_i)$, where $s_i$, $t_i$ represent the entity span and type of $y_i$, respectively. $s$ will have different formats according to the tagging schemes. In conventional generation-based methods for NER, the target sequence $Y = [(s_1, t_1), (s_2, t_2), ..., (s_m, t_m)]$ with $m$ entities will then be uniquely determined, following the strict \textit{left-to-right} order as in the source sequence. Then the generation procedure can be formulated as the following equation:
\begin{equation}
    P(Y|X) = \prod_{i=1}^{|Y|} P(y_i|X, Y_{<i})
\end{equation}

\subsection{Augment Data via Re-ordering}
A straightforward basis for re-ordering is position. Then augmented samples can be constructed by randomly shuffling~\cite{zhang-etal-2022-de-2}. However, there may not be a clear distinguishing criteria for these augmented entity sequences to further solve the one-to-many problem. So in our work, we alternatively choose the entity types as the principal factor of re-ordering.

$T=\{t_1, t_2, ..., t_l\}$ is the entity type set with cardinality $l$ for a certain dataset. We use $p$ to denote a random permutation of elements in $T$, such as [$t_3, t_1, t_2, ...$]. Following the specific entity type order $p$, the original target sequence $Y$ in the \textit{left-to-right} order can be re-ordered into $Y_p$ as:
\begin{equation}
    Y_p = [..., [(E_{p_i, 1}, p_i), ...,(E_{p_i, n_{p_i}}, p_i)], ...]
\end{equation}
where $E_{p_i, j}$ represent the $j$-th entity span with type $p_i$, following the original order, and $n_{p_i}$ indicates the number of entities with type $p_i$. Thus, in the sequence $Y_p$, there will be $l$ tuples like $[(E_{p_i, 1}, p_i), ...,(E_{p_i, n_{p_i}}, p_i)]$, and each represents a set of entities with a same type.

Given an entity type set T with cardinality $l$, $l!$ permutations like $p$ can be easily obtained. We denote them as a set $\mathrm{Perm}(T)$. For each $p \in \mathrm{Perm}(T)$, we can get a unique re-ordered $Y_p$.

As an example, the original entity sequence is $Y=$ ``[(\textbf{\textit{EU}}, {\color{red} MISC}), (\textbf{\textit{Britain}}, {\color{blue} LOC}), (\textbf{\textit{BSE}}, {\color{red} MISC})]''.
If the order $p$ is given as ``PER, LOC, MISC, ORG'', $Y$ will then be gathered as $Y_p=$ ``[[(\textbf{\textit{Britain}}, {\color{blue} LOC})], [(\textbf{\textit{EU}}, {\color{red} MISC}), (\textbf{\textit{BSE}}, {\color{red} MISC})]]''\footnote{``EU'' and ``BSE'' stand for the ``EU-wide'' and ``Bovine Spongiform Encephalopath'' respectively in Figure \ref{fig:model}.}.

\subsection{Prompt-based Order Instructions}
To fully use the augmented target sequences and prevent the one-to-many mapping problem, we separate these sequences by different $p$. As depicted in Figure \ref{fig:model}, we construct prompt-based order instructions as ``Following the order: $p$'' for augmented entity sequences. These prompts indicate to the model which entity type to focus on at a certain generation step. As a result, the predictions among different entity types are naturally modeled and the target sequences are uniquely determined with respect to the instructions. In this way, we can enlarge our training samples by $|\mathrm{Perm}(T)|$ times.


\section{Experiments}
To evaluate the effectiveness of PODA over generation-based methods, we conduct several experiments over two flat NER datasets and one nested NER dataset in several few-shot settings.

\paragraph{Datasets} For flat NER datasets, we choose CoNLL-2003~\cite{tjong2003introduction} and MIT-Movie~\cite{Liu2013QueryUE} from two different domains.
As for nested NER subtask, we conduct experiments on ACE-2005~\cite{doddington-etal-2004-automatic-2}. We randomly select 15\% samples from the MIT-Movie training set as the development set. For the experiments on ACE-2005, we use the same data split as~\cite{lu-roth-2015-joint-2}. 

\paragraph{Experimental Settings} To show that our approach can be generally applied to different generation-based methods, we use pure \textit{T5-base} and \textit{Flan-T5-base}~\cite{chung2022scaling} as our main backbones. And we utilize \textit{BART-base} when conducting experiments over BART-NER. We run T5 and Flan-T5 for 40 epochs and BART-NER for 200 epochs. To keep the stability of the few-shot setting, we set the batch size to 2/2/4/8 for the 5/10/20/50 shot settings, respectively. The learning rate of the Adam optimizer is set to 2e-5/5e-5.

In this work, we follow~\cite{ma-etal-2022-template-2} and focus on few-shot settings that only $K$ samples of each entity type from the training set are provided for training on a certain dataset. 
We conduct experiments in $K = \{5, 10, 20, 50\}$ settings 
and report the mean and deviation performance over three splits. In this work, we adopt the same sampling strategy as \citet{yang-katiyar-2020-simple-2}\footnote{As different sampling may affect the performance, we also conduct experiments on the splits released in \citet{huang-etal-2021-shot-2} and the reported results can be referred in the appendix.}. For CoNLL-2003, we use all the permutations of the entity type set as there are only 4 types. For datasets with more entity types such as ACE-2005, we randomly choose 20 different order instructions.

\begin{table}[t!]
	\centering
	\renewcommand\thetable{2}
	\setlength{\tabcolsep}{6pt} 
	\renewcommand{\arraystretch}{1} 
	\resizebox{0.95\linewidth}{!}
{
    \begin{tabular}{ll|cc}
        \toprule
           \multirow{1}{*}{\textbf{K-shot}}&\multirow{1}{*}{\textbf{Models}} & \textbf{CoNLL-2003} & \textbf{ACE-2005}\\
           \midrule
           \midrule
           \multirow{2.5}{*}{\textbf{K=10}}&BART-NER& 23.43  $\pm$  5.16 &	25.64 $\pm$ 1.56 \\  
           \cmidrule(lr){2-4}
           &PODA& \textbf{39.14  $\pm$  3.55}&	\textbf{34.30 $\pm$ 1.54}\\
           
           \midrule
           \multirow{2.5}{*}{\textbf{K=20}}&BART-NER& 38.87  $\pm$  1.94 &	35.58 $\pm$ 1.59\\ 
           \cmidrule(lr){2-4}
           &PODA& \textbf{53.11  $\pm$  4.33}& \textbf{42.18  $\pm$  0.80}	\\
           
           \midrule
           \multirow{2.5}{*}{\textbf{K=50}}&BART-NER& 56.99  $\pm$  1.84 &	49.17  $\pm$ 1.54\\  
           \cmidrule(lr){2-4}
           &PODA& \textbf{68.06 $\pm$ 1.45}& \textbf{56.24  $\pm$  0.76}	\\
            \bottomrule
        \end{tabular}
}
	\vspace{-2mm}
	\caption{The performance of BART-NER and PODA (BART-NER) on two datasets with (K=10, 20, 50)-shot.}
	\label{tab:bart}
	\vspace{-3mm}
\end{table}

\subsection{Main Results}
In the experiments, we compare our method with several strong baselines  and competitive few-shot approaches. 

\textbf{NNShot} and \textbf{StructShot}~\cite{yang-katiyar-2020-simple-2} are two metric-based approaches. \textbf{Template-NER}~\cite{cui-etal-2021-template-2} constructs a template for a single entity type, and enumerates each span together with this type and calculates its generation probability. \textbf{Ent-LM}~\cite{ma-etal-2022-template-2} proposes a template-free prompt tuning method and induces the language models to predict label words at entity positions during fine-tuning, while \textbf{Ent-LM + struct} leverages the viterbi algorithm proposed
in \citet{yang-katiyar-2020-simple-2} to further boost the performance.

Table \ref{tab:mainresult} shows the results of our proposed PODA with these baselines. 
We only report the results of our method and the backbone model on experiments over ACE-2005 since traditional few-shot methods are hard to be applied to nested NER. 

Based on the results, we have the following observations: 
(1) For nearly all few-shot settings, our proposed method performs consistently better than the strong baselines. 
(2) It is worth noting that our method can outperform the backbone model T5 by 22.54/8.97/7.71 points when conducting 5-shot setting over these three datasets, which means generative PLMs like T5 are suffering from low-resource tuning and our method shows the strong ability of improving their training under few-shot settings. Even without the prompt-based order instructions, the augmented data can help the model achieve much better performance compared with pure T5, which demonstrates our claim that there is no need to keep the strict \textit{left-to-right} order. 
(3) By comparing with Template-NER~\cite{cui-etal-2021-template-2}, which is also a template-based prompt method, the results show the advantages of our method over traditional template-based prompt method. We fairly query all the entity types rather than constructing a template for each type.
(4) Regarding the experimental results on T5 and Flan-T5, it was observed that without our method, Flan-T5 consistently outperformed T5. However, our method demonstrated the ability to enhance the performance of both models in almost all scenarios, albeit with a relatively smaller improvement observed on Flan-T5. We suspect that this discrepancy is due to the disparity between our constructed data and natural language. While more enhanced data samples are provided (i.e. on MIT-Movie 50-shot setting), Flan-T5, which underwent instruction tuning, may become perplexed, leading to a relatively inferior performance compared to the pure T5 model. 

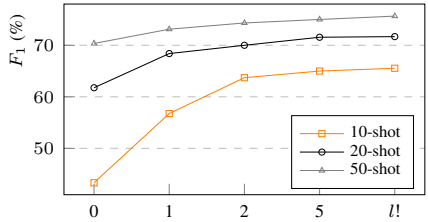
\begin{figure}
\centering
\resizebox{0.75\linewidth}{!}{
\begin{tikzpicture}
\pgfplotsset{width = 6.5cm, height = 4.2cm}
    \begin{axis}[
        ylabel={$F_1$ (\%)},
        ymax=78,
        ymin=41,
        y label style={yshift=-0.6cm, xshift=0.8cm},
        x label style={yshift=0.4cm, xshift=2.8cm},
        label style={font=\fontsize{7}{1}\selectfont},
        xtick = {1,2,3,4,5},
        xticklabels = {0, 1, 2, 5, $l!$},
        xticklabel style = {font=\fontsize{7}{1}\selectfont, rotate=0,},
        yticklabel style = {font=\fontsize{7}{1}\selectfont},
        xtick pos = left,
        ytick pos = left,
        legend pos = south east,
        legend style={font=\fontsize{6.5}{1}\selectfont, row sep=-0.1cm,/tikz/every odd column/.append style={column sep=0.01cm}},
        ymajorgrids = true,
        grid style=dashed,
    ]
    \addplot [mark=square, mark size=1.2pt, color=orange] plot coordinates {
    (1, 43.32) (2, 56.75) (3, 63.71) (4, 65.00) (5, 65.54)};
    \addlegendentry{10-shot};
    \addplot [mark=o,  mark size=1.2pt] plot coordinates {
    (1, 61.76) (2, 68.39) (3, 70.00) (4, 71.56) (5, 71.68)};
    \addlegendentry{20-shot};
    \addplot [mark=triangle,  mark size=1.2pt, color= gray] plot coordinates {
    (1, 70.35) (2, 73.12) (3, 74.33) (4, 75.01) (5, 75.66)};
    \addlegendentry{50-shot};
    \end{axis}
\end{tikzpicture}
}
\vspace{-2mm}
\caption{The performance of PODA (T5) on CoNLL-2003 with different number of permutations, where \textit{0} and \textit{$l!$} means training with no and all permutations.}
\vspace{-3mm}
\label{fig:type}
\end{figure}

\subsection{Analysis}
\paragraph{Generalization Ability} 
Our method can also be applied to other generative backbones, such as UIE~\cite{lu-etal-2022-unified-2} and BART-NER~\cite{yan-etal-2021-unified-generative-2}. To test the generalization ability and simplify the generation procedure, we test our approach over BART-NER since the target sequences of UIE include extra information. In order to alleviate the poor-tuning problem, we run BART-NER for 200 epochs. We only include the results of pure and enhanced (with PODA) BART-NER in Table \ref{tab:bart} for clear comparison. The results show that BART-NER will have low-resource tuning even worse than pure T5 model, and our method can uniformly help BART-NER achieve a reasonable performance. In addition, we also have an interesting observation. By comparing results between Table \ref{tab:mainresult} and \ref{tab:bart}, the results of BART-NER are uniformly lower than those of T5 on CoNLL-2003. But on ACE-2005, although the results are lower in 10-shot setting, BART-NER slightly performs better than T5 in 20-shot and outperforms T5 by 6.07 points in 50-shot. It is possible that the tagging schema of BART-NER is more compatible with nested NER. Our method can still further improves its performance, which means PODA has the ability to be uniformly generalized to different tagging schemes. 
\paragraph{Improvements with More Permutations}
As illustrated in the experiment settings, we use all the permutations to construct the order instructions since there are only 4 entity types in CoNLL-2003. To see whether the augmented data is valid, we also test the performance with increasing numbers of permutations. As visualized in Figure \ref{fig:type}, we can observe significant performance improvement with only one re-ordered target sequence, and the model can be further improved with increasing permutations, which also demonstrates the effectiveness of our augmented data.

\section{Conclusions}
In this paper, we propose PODA to improve the training of various generation-based NER methods (e.g. T5 and BART-NER) in different few-shot settings. By eliminating the strict left-to-right order assumption in traditional generation-based NER methods, PODA can construct sufficient while reasonable target entity sequences, thus leading to improved model training. To address situations where a single source sequence may have multiple target sequences, we additionally propose order instructions to facilitate the disambiguation of this one-to-many mapping. Experimental results demonstrate the effectiveness and generalization capability of both our data augmentation method and the prompt-based order instructions.

\section*{Limitations}
Although our approach can improve the training of generation-based NER methods, there are still some limitations and we leave as future directions to explore.
\paragraph{More Diverse Decoding Strategy} In the training stage of our method, we concatenate the input sentence with all different prompt-based order instructions. But when evaluating, we only use the instruction ``from left to right''. We believe that if there is a proper algorithm that can select entities inside all the target entity sequences generated with different instructions, the performance will be further improved.
\paragraph{Different Prompt Design} As shown in previous work~\cite{Sanh2021MultitaskPT}, different prompts may affect the performance. In our work, we utilize some straightforward prompts as order instructions rather than specially designed. Using some special tokens in PLMs may also be helpful and a controllable generation-style method will then be proposed.

\bibliography{anthology,custom}
\bibliographystyle{acl_natbib}

\clearpage

\end{document}